\documentclass[letterpaper, 10 pt, conference]{ieeeconf}  
\usepackage[left=54pt, right=54.5pt, top=54.5pt, bottom=57pt]{geometry}

\IEEEoverridecommandlockouts                              

\usepackage{graphicx}
\usepackage{subcaption}
\usepackage{color}

\title{\LARGE \bf
Deep Imitation Learning for Autonomous Driving in Generic Urban Scenarios with Enhanced Safety
}

\author{Jianyu Chen*, Bodi Yuan* and Masayoshi Tomizuka
\thanks{* indicates equal contribution}
\thanks{J. Chen, B. Yuan and M. Tomizuka are with Department of Mechanical Engineering, University of California, Berkeley, CA94720, USA. \quad Corresponding to {\tt\small jianyuchen@berkeley.edu}}%
\thanks{This work was supported by Denso International at America}
}

\begin{document}

\maketitle
\thispagestyle{empty}
\pagestyle{empty}

\begin{abstract}

The decision and planning system for autonomous driving in urban environments is hard to design. Most current methods manually design the driving policy, which can be expensive to develop and maintain at scale. Instead, with imitation learning we only need to collect data and the computer will learn and improve the driving policy automatically. However, existing imitation learning methods for autonomous driving are hardly performing well for complex urban scenarios. Moreover, the safety is not guaranteed when we use a deep neural network policy. In this paper, we proposed a framework to learn the driving policy in urban scenarios efficiently given offline connected driving data, with a safety controller incorporated to guarantee safety at test time. The experiments show that our method can achieve high performance in realistic simulations of urban driving scenarios.

\end{abstract}

\section{INTRODUCTION}
Decision and planning for autonomous driving in urban scenarios with dense surrounding dynamic objects is particularly challenging. The difficulties come from multiple respects: 1) Complex road conditions including different road topology, geometry and road markings in various scenarios such as intersection and roundabout; 2) Complex multi-agent interactions where their coupled future motions are unknown; 3) Various traffic rules such as traffic lights and speed limit.

The majority of autonomous driving community, including both industry and academy, is focusing on the non-learning model-based approach for decision making and planning. This model-based approach often requires manually designing the driving policy model. For example, a popular pipeline is to first do scenario-based high-level decision, then predict the future trajectory of surrounding objects, and then treat the predicted trajectories as obstacles and apply motion planning techniques to plan a trajectory for the ego vehicle~\cite{thrun2006stanley,montemerlo2008junior,paden2016survey,gonzalez2016review}. 

However, the manually designed policy model is often sub-optimal. There are two main reasons for this: 1) The model-based approach often requires defining some motion heuristics or at least some cost functions to indicate what does a desired decision and planning look like. However, designing an accurate cost function that can make the vehicle do what we \textit{really} want can be extremely difficult~\cite{hadfield2017inverse}; 2) For highly entangled interactions among multiple agents, simple policy models are not adequate. However, complex behavior models such as game theoretic models are not solvable in their general form (e.g, general sum multi-player games) with the current non-learning methods. Besides its sub-optimality, the model-based approach is also expensive with respect to development and maintenance, as it relies on human engineers to improve its performance.

While it is difficult to design a decision and planning system for autonomous driving, an experienced human driver can solve the driving problem easily, even in extremely challenging urban scenarios. Thus an alternative is to learn a driving policy from human driver experts using imitation learning. Applying imitation learning has several benefits. First, we do not need to manually design the policy model or the cost function which can be sub-optimal. Second, we only need to provide expert driving data which is not difficult to obtain at scale, and the computer will then learn a driving policy automatically.

There are already existing works of imitation learning approaches for driving which typically focus on predicting direct control commands such as steering and braking from raw sensor data such as camera images~\cite{bojarski2016end,dosovitskiy2017carla,codevilla2018end,sauer2018conditional}. However they can only handle simple driving tasks such as lane following. Direct mapping from raw sensor data to control is too complex, which requires a huge amount of training data to cover most situations. Besides, the end-to-end architecture lacks transparency as we cannot explain the decision making process in a neural network, which makes it hard to evaluate and debug.

A more serious problem for the current imitation learning approaches, especially with function approximation such as deep neural network, is safety. Currently no theoretical results can guarantee the safety of a policy composed of a deep neural network. Safety is the most crucial issue for autonomous driving and it must be considered strictly.

In this paper, we propose a framework which efficiently obtains the intelligence of decision making for handling complex urban scenarios using imitation learning, and then provides safety enhancement to the learned deep neural network policy. We design a bird-view representation as the input and define future trajectory as the output of the driving policy, which is similar to Waymo \cite{waymo} and Uber's recent works \cite{uber}. The designed representation significantly reduces the sample complexity for imitation learning. Then a safety controller based on safe set theory is incorporated, which generates control commands to track the planned trajectory while guaranteeing safety. Experiments show that the framework is able to obtain a deep convolutional neural network policy which is intelligent enough to achieve high performance in generic urban driving scenarios, with only 100k training examples and 20 hours training time on a single GTX 1080 Ti. 

\section{RELATED WORKS}
The first imitation learning algorithm applied to autonomous driving was 30 years ago, when ALVINN system~\cite{pomerleau1989alvinn} used a 3-layer neural network to perform road following based on front camera images. Helped by recent progress in deep learning, NIVIDIA developed an end-to-end driving system using deep convolutional neural networks~\cite{bojarski2016end,bojarski2017explaining}, which can perform good lane following behaviors even in challenging environments where no lane markings can be recognized. Researchers also trained deep neural networks to predict the control output from camera image and evaluate their open loop performance (e.g, the prediction error). \cite{xu2017end} used an FCN-LSTM architecture with a segmentation mask to train a deep driving policy. \cite{wang2018deep} proposed an object-centric model to predict the vehicle action with higher accuracy. Although both \cite{xu2017end} and \cite{wang2018deep} achieved good prediction performance for complex urban scenarios, they did not provide closed loop evaluation either on real world or simulated environments. \cite{sun2018fast} used imitation learning to drive a simulated vehicle in closed loop, however it is restricted to limited scenarios such as lane following and lane changing with fixed number of surrounding vehicles.

CARLA simulator~\cite{dosovitskiy2017carla} has been developed and open-sourced recently. It enables training and testing autonomous driving systems in a realistic three-dimensional urban driving simulation environment. Based on CARLA, \cite{codevilla2018end} used conditional imitation learning to learn an end-to-end deep policy that follows high level commands such as go straight and turn left/right. \cite{sauer2018conditional} defined several intermediate affordance such as distance to objects, learned a deep neural network to map camera image to the affordance, and then performed model-based control based on the affordance.

The above methods are all using front camera images as the input. However the complexity of direct visual information has limited the performance of such methods. Bird-view representation is a good way to simplify the visual information while maintaining useful information for driving. Uber~\cite{uber,cui2018multimodal} used a rasterized image which includes information for the map and objects as the input, and learned a convolutional neural network to predict the future trajectory of the vehicle. Waymo~\cite{waymo} used a similar mid-to-mid representation and learned a deep model that combined with perception and control modules, could drive a vehicle through several urban scenarios.
 
Collision avoidance safety is an important topic in robotics. Potential field method~\cite{koren1991potential} introduces an artificial field around the obstacle and push the vehicle away when the distance is close. The method is efficient, but it cannot guarantee the safety. Reachability analysis~\cite{mitchell2005time} calculates the reachable set of the agent's state using game theory, and constrains the agent from reaching unsafe state. However, it is computationally expensive. Planning based methods~\cite{chen2018foad,chen2017constrained} may achieve safe motion in real time computation, but it requires the prediction of trajectories for each object. In this work we use safe set algorithm~\cite{liu2017robustly,liu2016enabling} to develop our safety controller, which is guaranteed to be safe, computationally efficient and do not require prediction of obstacles' future motion.

\vspace{-0.5mm}

\section{FRAMEWORK OVERVIEW}
\begin{figure*}
\centering
  \includegraphics[width = .69\textwidth]{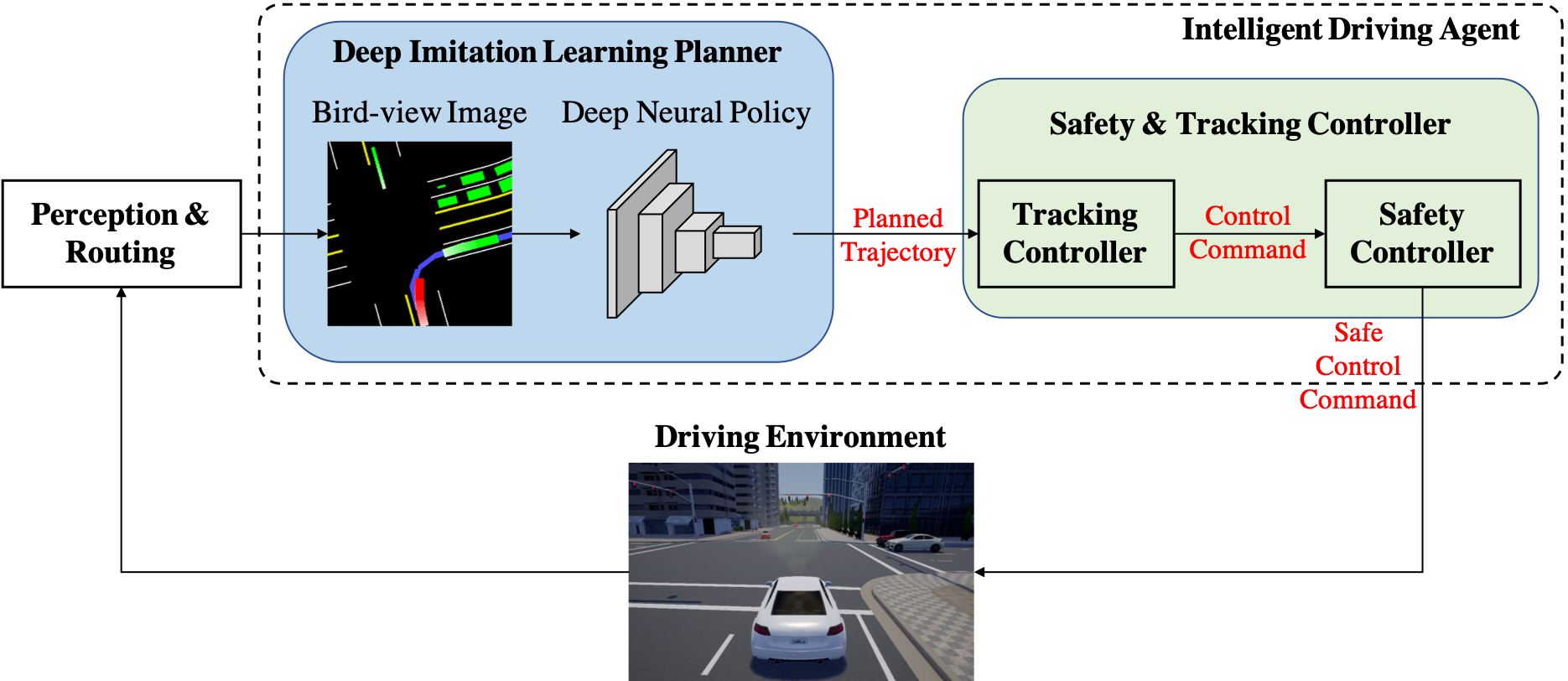}
  \caption{\label{Fig:framework}\em{Framework overview of the our system. The agent takes information from the perception and routing modules, generates a bird-view image and outputs the planned trajectory using a deep neural policy. The safety \& tracking controller then calculates the safe control command to be applied to the ego vehicle in the driving environment.}}
  \vspace{-6mm}
\end{figure*}

Our system acts as an intelligent driving agent in a closed loop environment, as shown in Fig.\ref{Fig:framework}. The agent receives routing and perception information from the driving environment. It then outputs the control command such as throttle, steering and braking to be applied to the ego vehicle. The system includes two main building blocks: a deep imitation learning trajectory planner and a safety \& tracking controller. The deep imitation learning trajectory planner is responsible to handle almost everything about driving intelligence, such as how to follow the given route in various road conditions, how to react to surrounding objects, and how to handle different traffic light states. This module is learned end-to-end from the perception results to the planned trajectory. The safety \& tracking controller is responsible to guarantee safety and handle vehicle dynamics. This module is designed with non-machine-learning methods.

In this work, we assume that we already have a functioning perception module to process the raw sensor data. For example, we can use a localization system to estimate ego vehicle pose, an object detection system to detect the bounding box, position and heading of surrounding objects, and a traffic light detector to tell the states of traffic lights. Furthermore, we have access to the High-Definition map data for the area that the ego vehicle is operating, as well as the routing information which guides the ego vehicle to a specified goal position. This is a reasonable assumption as they are not difficult to obtain with current autonomous driving technology, at least under a good weather. We will not discuss the development and properties of the perception module in this work, but concentrate on the decision and planning part.

\section{DEEP IMITATION LEARNING FOR URBAN AUTONOMOUS DRIVING}
The goal of imitation learning is to learn a controller that imitates the behavior of the expert. At data collection phase, an expert (either a human driver or a controller) receives observation ${\bf{o}}_t$ and output action ${\bf{a}}_t$ at time step $t$. The observation-action pairs ${\cal D} = \left\{ {\left( {{{\bf{o}}_t},{{\bf{a}}_t}} \right)} \right\}_{t = 1}^N$ are then stored as the dataset. Let the policy function be $f\left( {{\bf{o}};\theta } \right)$, where $\theta$ is its parameter. $f$ can be any function approximator, including deep neural network as used in this paper. The imitation learning problem is then formulated as a supervised learning problem, where the goal is the optimize the policy function parameter $\theta$ to minimize the loss function ${\cal L}$: 

\vspace{-2mm}
\begin{equation}\label{Eq:sl}
    \mathop {\min }\limits_\theta  \sum\limits_{\left( {{{\bf{o}}_i},{{\bf{a}}_i}} \right) \in {\cal D}} {{\cal L}\left( {f\left( {{{\bf{o}}_i};\theta } \right),{{\bf{a}}_i}} \right)} 
\end{equation} 
\vspace{-2mm}

In this section, we will introduce how we design observation $\bf{o}$, action $\bf{a}$, policy function $f$, loss function ${\cal L}$, and how we augment the data to obtain a robust policy.

\subsection{Observation-action Representation}
A straight forward input-output representation is raw sensor data (e.g, front view camera image) for observation, and direct control command (e.g, throttle, steering, braking) for action. However, directly learning the complex mapping from raw sensor data to control output is too inefficient and hard to generalize. The raw sensor data contains extremely high dimensional information which can be influenced by different textures and appearances of roads and objects, different weather conditions, and different daytime. To allow generalization of the learned policy, the dataset needs to cover enough data for each aspect of sensor information such as texture, weather, light condition and object appearance. The direct control output is also influenced by different vehicle dynamics, thus a new policy needs to be trained if the vehicle dynamics is changed. 

\begin{figure}
    \centering
    \includegraphics[width = .43\textwidth]{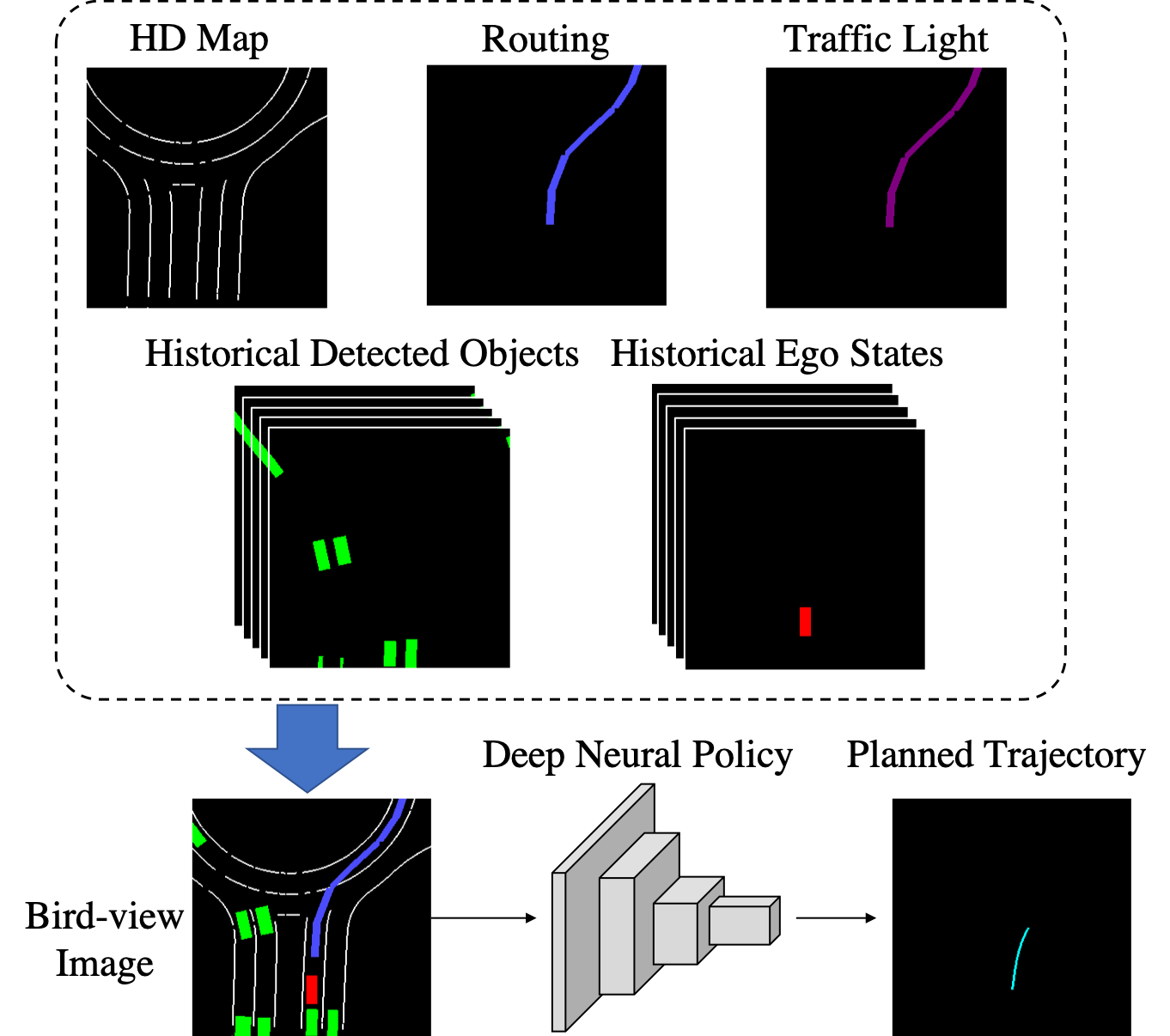}
    \caption{\em{Observation-action Representation of our deep imitation learning planner. The bird-view observation combines information of HD map, routing, traffic light, historical detected objects and historical ego states. The output action is a planned trajectory represented by a vector.}}
    \label{Fig:representation}
    \vspace{-6mm}
\end{figure}

To alleviate this problem, we used a bird-view representation for the observation ${\bf{o}}_t$. This bird-view representation is a concise description of only the useful information for decision making and planning, discarding irrelevant information such as texture. As shown in Fig.\ref{Fig:representation}, our bird-view input representation is composed of the following five parts:

\subsubsection{\bf{High-definition (HD) Map}}
The HD map contains information of road conditions. Here we render all the lane markings represented by yellow or white polylines on the 2D RGB image.

\subsubsection{\bf{Routing}}
The routing information is provided by a route planner after we define the start and end point. It is represented by a sequence of waypoints for the ego vehicle to follow. We render it as a thick blue polyline
in the image.

\subsubsection{\bf{Traffic Light State}}
When the state of the traffic light becomes red, we set the color of the route to be purple, otherwise it will maintain blue.

\subsubsection{\bf{Past Detected Objects}}
The past detected surrounding objects (e.g, vehicles, bicycles, motorcycles) in a past time period are rendered as green boxes, with reduced level of brightness meaning earlier time-steps.

\subsubsection{\bf{Past Ego States}}
Similar to the detected objects, the past ego states are represented as boxes with reduced brightness. The color of the boxes are set red. 

\vspace{2mm}

The final bird-view image is rendered with pixel size $192 \times 192$, and is always aligned to the ego vehicle's local coordinate. The actual size of the field of view is $\left( {40m,40m} \right)$, where the ego vehicle is positioned at $\left( {20m,8m} \right)$.

The action of the policy is also altered from direct control output to trajectory ${\bf{a}}_t = \left[ {{x_{t + 1}},{y_{t + 1}}, \cdots ,{x_{t + H}},{y_{t + H}}} \right]$, where $x_i$ and $y_i$ are $x$ and $y$ position in the local coordinate of ego vehicle at time step $i$, $H$ is the preview horizon and $t$ is the current time step. While direct control output can be significantly influenced by vehicle dynamics, the future trajectory would have little difference if the vehicle dynamics does not change too much. The output trajectory can be tracked by a vehicle specific tracking controller, which is easy to design as will be illustrated in the next section.

\subsection{Network architecture}
Taking the bird-view image as input, we use a CNN model to predict ego vehicle's future trajectory. Our CNN model has the same conv-layers as VGGNet16~\cite{vggHe2016deep}, followed by a fully connected layer with 1000 hidden units, which is then fully connected with the final output layer with $2H$ units. The output layer represents the $H$ predicted trajectory points $(\hat{x}_{t+i}, \hat{y}_{t+i})$ in the ego vehicle local coordinate.

We want to optimize the displacement error $d_{t+i}$ between the expert's actual trajectory point position $({x}_{t+i}, {y}_{t+i})$ and the predicted point position $(\hat{x}_{t+i}, \hat{y}_{t+i})$:
\[d_{t+i} = (({x}_{t+i}-\hat{x}_{t+i})^2 + ({y}_{t+i}-\hat{y}_{t+i})^2)^{\frac{1}{2}} . \]
The overall loss function is defined as 
\[ {\cal L}_t = \frac{1}{H} \sum_{i=1}^{H} d^2_{t+i} .\]

\subsection{Data augmentation}
If we directly solve the supervised learning problem (\ref{Eq:sl}), the resulting policy may be unstable and the vehicle will easily run out of the road. This is due to the co-variant shift of the vanilla imitation learning algorithm, which only learns from normal driving data. In the test phase, small prediction error can be accumulated and the vehicle may reach some unseen states so that it is unable to recover.

To solve this problem, we periodically introduce noise to the expert controller during the data collection phase, and let the expert recover from the perturbation. The control noise is added every 8 seconds, and will last for 1 second. The vehicle's pose might be pushed away from the waypoints. The expert then provides demonstrations of recovering from perturbations. The states during the noise phase are removed in order not to contaminate the dataset. This data augmentation trick significantly improves the performance of the learned policy, as shown in our experiments.

\section{SAFETY ENHANCEMENT \& TRAJECTORY TRACKING CONTROL}
Since we use deep neural network, the safety and feasibility of the planned trajectory cannot be guaranteed. In this section, the design of the safety enhancement controller and the trajectory tracking controller will be introduced.

\subsection{Trajectory Tracking Controller}
Given the planned future trajectory $\left[ {{\hat{x}_{t + 1}},{\hat{y}_{t + 1}}, \cdots ,{\hat{x}_{t + H}},{\hat{y}_{t + H}}} \right]$, a tracking controller is implemented to calculate the desired acceleration $a_t$ and steering angle $\delta_t$ to drive the vehicle that follows the trajectory. A target waypoint $\left( {{{\hat x}_{t + m}},{{\hat y}_{t + m}}} \right)$ is selected where $1 \le m \le H - 1$ ($m=5$ in this paper). The controller is then decoupled to longitudinal and lateral control:

\subsubsection{Longitudinal Controller}
The target speed is set to be 
\[{v_d} = \frac{1}{{dt}}{\left\| {\left( {{{\hat x}_{t + m + 1}},{{\hat y}_{t + m + 1}}} \right) - \left( {{{\hat x}_{t + m}},{{\hat y}_{t + m}}} \right)} \right\|_2}\]
where $dt$ is the time interval between two consecutive time steps. The desired acceleration $a_t$ is then obtained using PID control to eliminate the speed tracking error ${e_v}\left( t \right) = {v_d} - v\left( t \right)$, where $v\left( t \right)$ is the current speed of ego vehicle.

\subsubsection{Lateral Controller}
The normalized vector from the ego vehicle position to the target way point is ${{\bf{n}}_{{\rm{target}}}} = \frac{{\left( {{{\hat x}_{t + m}},{{\hat y}_{t + m}}} \right)}}{{{{\left\| {\left( {{{\hat x}_{t + m}},{{\hat y}_{t + m}}} \right)} \right\|}_2}}}$. The normalized vector of the ego vehicle heading is ${{\bf{n}}_{ego}}\left( t \right) = \left( {\cos {\theta _t},\sin {\theta _t}} \right)$, where $\theta_t$ is the yaw angle of the ego vehicle. Then the desired steering angle is obtained using PID control to eliminate the heading error:
\[{e_{yaw}}\left( t \right) = {\cos ^{ - 1}}\left( {{{\bf{n}}_{ego}}\left( t \right) \cdot {{\bf{n}}_{target}}\left( t \right)} \right)\]

\subsection{Safety Enhancement Controller}
The acceleration and steering command $a_t$ and $\delta_t$ calculated by the tracking controller does not guarantee safety. We incorporate a safety controller that will modify $a_t$ and $\delta_t$ to enhance safety, if their original values are not safe.

\begin{figure}
    \centering
    \includegraphics[width = .37\textwidth]{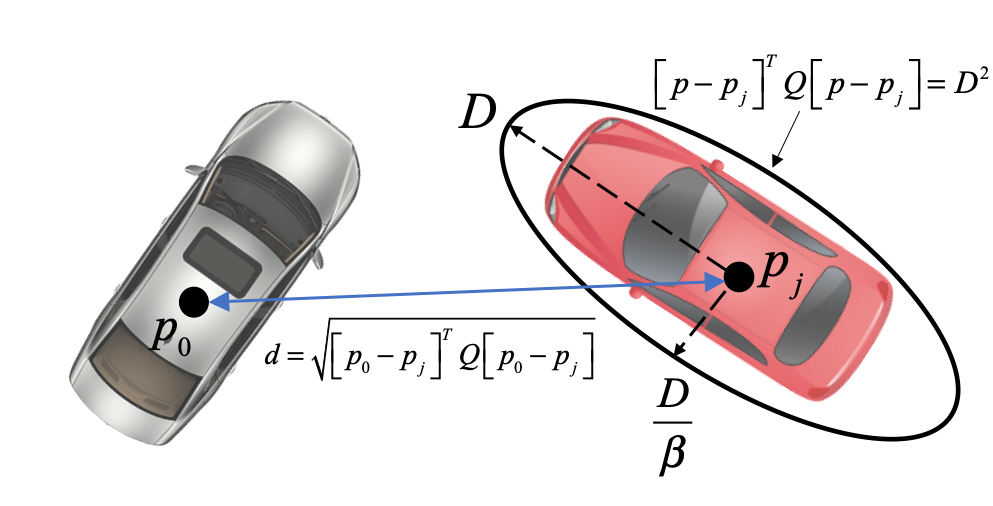}
    \caption{\em{Illustration of the safety index. Gray is the ego vehicle, red is a surrounding vehicle. The safety constraint is similar to the ellipse around the red vehicle, while also considering the relative speed of the two vehicles.}}
    \label{Fig:safety}
    \vspace{-5mm}
\end{figure}

Our method is developed based on the safe set algorithm~\cite{liu2017robustly,liu2016enabling}. The key idea is that for each time step $t$, we will calculate a control safe set ${U_S}\left( t \right)$ of the control command $u\left( t \right) = {\left[ {\begin{array}{*{20}{c}}
{{a_t}}&{{\delta _t}}
\end{array}} \right]^T}$. The control safe set has the property that if $u\left( t \right) \in {U_S}\left( t \right)$, the ego vehicle will stay safe.

To obtain ${U_S}\left( t \right)$, a definition of safety needs to be stated. Here we define a safety index $\phi \left( x \right)$, which is a function of the state $x$, where $x$ represents states (e.g, position, velocity, heading) of both the ego vehicle and a surrounding object. In this paper, the safety index is defined as:

\[\phi \left( x \right) = D - {d^2}\left( x \right) - \alpha \dot d\left( x \right)\]
where $d\left( x \right)$ is a shaped distance between the ego vehicle and the surrounding vehicle:

\[d\left( x \right) = \sqrt {{{\left[ {{p_0} - {p_j}} \right]}^T}Q\left[ {{p_0} - {p_j}} \right]} \]
where $p_0$ indicates the position of the ego vehicle and $p_j$ indicates the position of the surrounding vehicle. $Q$ is a 2-by-2 matrix such that ${\left[ {{p} - {p_j}} \right]^T}Q\left[ {{p} - {p_j}} \right] = 1$ represents an ellipse around the surrounding vehicle with long axis equal to 1 and short axis equal to $\frac{1}{\beta }$, where $\beta$ is the aspect ratio of the ellipse. Let the state safe set $X_S$ be the level set of the safety index ${X_S} = \left\{ {x:\phi \left( x \right) \le 0} \right\}$. Then intuitively, the state safe set introduces an ellipse constraint as shown in Fig.\ref{Fig:safety}. It also considers the relative speed between the ego and surrounding vehicle. If their relative speed is high, it is more likely to be unsafe.

We can choose the control safe set to be ${U_S}\left( t \right) = \left\{ {u\left( t \right):\dot \phi  \le -\eta \,\,{\rm{if}}\;\phi  \ge 0} \right\}$ where $\eta  > 0$ is some margin. It can be proved that if $x\left(0\right) \in X_S$ and $u\left(t\right) \in U_S$ for $t \ge 0$, then $x\left(t\right) \in X_S$. Now if we approximate the ego vehicle dynamics to a control affine function $\dot x = f\left( x \right) + Bu$, the control safe set can be written as:

\vspace{-2mm}
\[{U_S}\left( t \right) = \left\{ {u\left( t \right):L\left( t \right)u\left( t \right) \le S\left( t \right)\,\,{\rm{if}}\;\phi  \ge 0} \right\}\]
where $L\left( t \right) = \frac{{\partial {x_0}}}{{\partial {x_j}}}B$ and $S\left( t \right) =  - \eta  - \frac{{\partial \phi }}{{\partial {x_j}}}{{\dot x}_j} - \frac{{\partial \phi }}{{\partial {x_0}}}f$, $x_0$ and $x_j$ are the states of the ego and surrounding vehicle, respectively. 

\begin{figure}
    \centering
    \includegraphics[width = .43\textwidth]{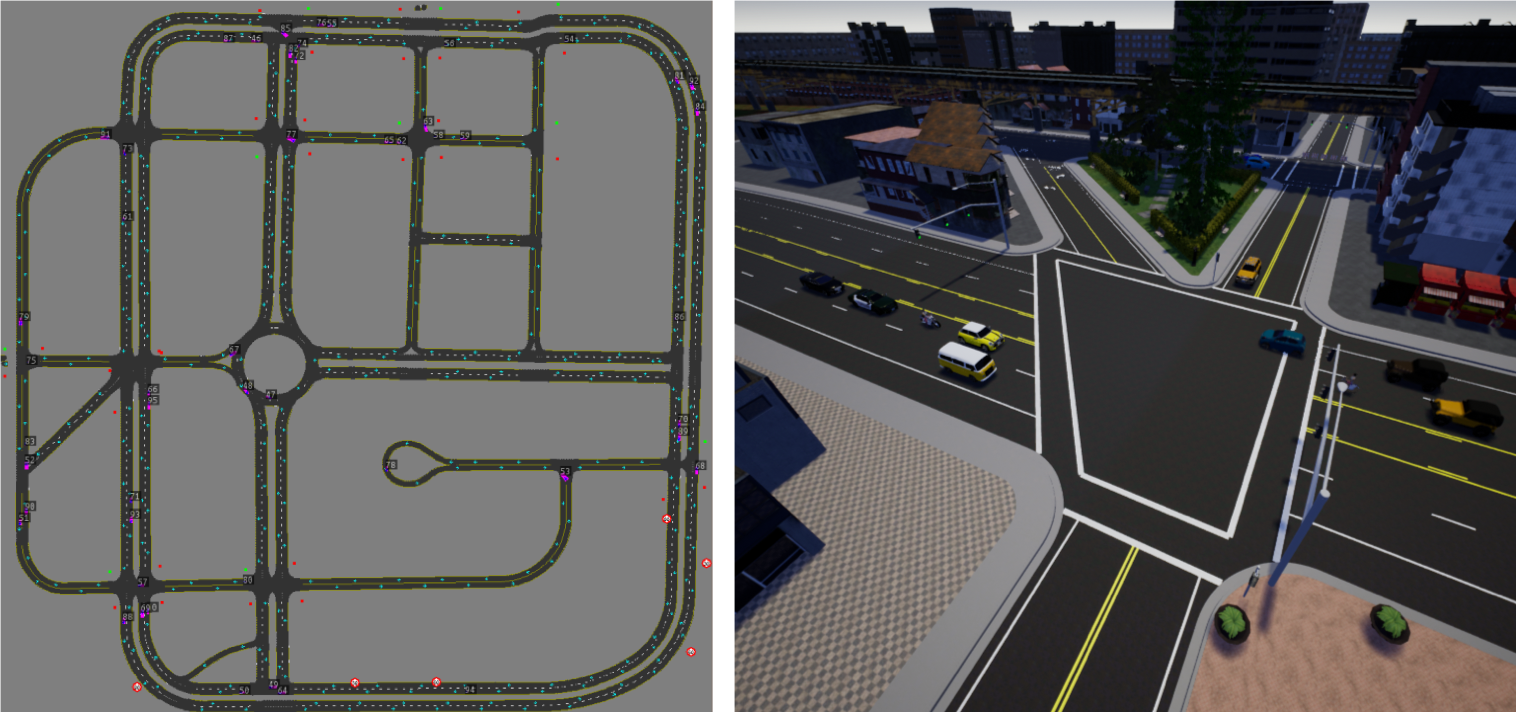}
    \caption{\em{The simulation environment we use. Left is the map layout, right is a sample view at an intersection.}}
    \label{Fig:carla}
    \vspace{-5mm}
\end{figure}

If there are multiple surrounding objects, we can calculate the intersection of the control safe set for each object, which is a convex polytope. Letting $u\left( t \right) = {\left[ {\begin{array}{*{20}{c}}
{{a_t}}&{{\delta _t}}
\end{array}} \right]^T}$ denotes the control command output from the trajectory tracking controller, the safety controller maps it into the control safe set $U_S$ by solving the following quadratic programming problem:

\vspace{-1mm}
\[{u^ * }\left( t \right) = \mathop {\arg \min }\limits_{u \in {U_S}} \frac{1}{2}{\left( {u - u\left( t \right)} \right)^T}W\left( {u - u\left( t \right)} \right)\]
where $W$ is a 2-by-2 weight matrix. We can thus obtain the modified safe control command $u^*\left( t \right) = {\left[ {\begin{array}{*{20}{c}}
{{a^*_t}}&{{\delta^* _t}}
\end{array}} \right]^T}$. Then the low-level controller will track the given acceleration $a^*_t$ and steering angle $\delta^*_t$.

Note that our safety controller is not restricted to be applied together with the specific imitation learning planner in this paper. It can be applied as a module with any upper level planner to modify their control output to enhance safety.

\section{EXPERIMENTS}
\subsection{Simulation Environment and Data Collection}
We collect data and evaluate our proposed method on CARLA simulator~\cite{dosovitskiy2017carla}. CARLA is an open-source high-resolution simulation platform for development and validation of autonomous driving systems. It simulates not only the raw sensor data such as camera image and Lidar point cloud, but also detailed vehicle dynamics. A system evaluated on CARLA is likely to have similar performance if applied to a real driving environment. Furthermore, in our system we use the processed bird-view image as input, which has no domain difference with that of the real world and thus the policy can be easily transferred from simulation to real world.

Fig.\ref{Fig:carla} shows the map layout and a sample view of the simulation environment we use for training. It includes various urban scenarios such as intersection and roundabout. The map has a range of $400m \times 400m$, containing about $6km$ total length of roads. We put 100 vehicles running autonomously in the simulator to simulate a multi-agent environment. The vehicles will randomly choose a direction at the intersection, follow the route, slow down for front vehicles and stop when the traffic light is red. 

At data collection phase, we use a model-based controller to act as the expert. The controller is the same as other agents. When ego vehicle is running, we record the rendered bird-view image and the corresponding ego vehicle state (global positions and yaw angle) every 0.1 second. The future ego vehicle trajectory is calculated by transforming its future global positions to the ego vehicle's current local coordinate.

\subsection{Bird-view Image Generation}
To render the bird-view input image, we build a buffer to store the historical states (position, velocity, heading, size) of all vehicles. The states are then transformed to the ego vehicle's current local coordinate. The HD map contains information of lane markings, which is extracted from the OpenDrive data provided by CARLA. Routing information is a sequence of waypoints provided by the global planner of CARLA, and is rendered as a thick blue line.

\subsection{Training}
We run the simulation for about 5 hours and generated 120k frames. 100k frames are used for training and 20k for evaluation. The model is trained from scratch using Adam optimizer \cite{kingma2014adam}, with initial learning rate of $10^{-4}$ for 30 epochs. It is then fined-tuned with learning rate of $10^{-5}$ for another 10 epochs. Batch size is set to 50. The model converges in about 20 hours on a single GTX 1080 Ti. 

\subsection{Models}
Besides our final model with data augmentation and safety controller, we also train and test the models without data augmentation and/or without safety controller for comparison. We thus have three models: 1) $M_0$ - the model without data augmentation and safety controller; 2) $M_1$ - the model with data augmentation but without safety controller; 3) $M_2$ - the model with both data augmentation and safety controller.

\subsection{Open Loop Evaluation}
For open loop evaluation, we calculate the average displacement error in both Town03 (Training Condition) and Town01 (New Town), as shown in Table \ref{OpenLoop}.
  
\begin{table}
\centering
\caption{\em{Average prediction displacement error (in meters)}}
\resizebox{0.35\textwidth}{!}{
\begin{tabular}{ccc}
\hline
        & Training Condition & New Town\\ \hline
$M_0$ &  0.16              &      0.44       \\ 
$M_1$ &  0.18              &      0.29       \\ \hline
\end{tabular}}
\label{OpenLoop}
\vspace{-5mm}
\end{table}

\begin{figure}
  \centering
	\includegraphics[width=0.46\textwidth]{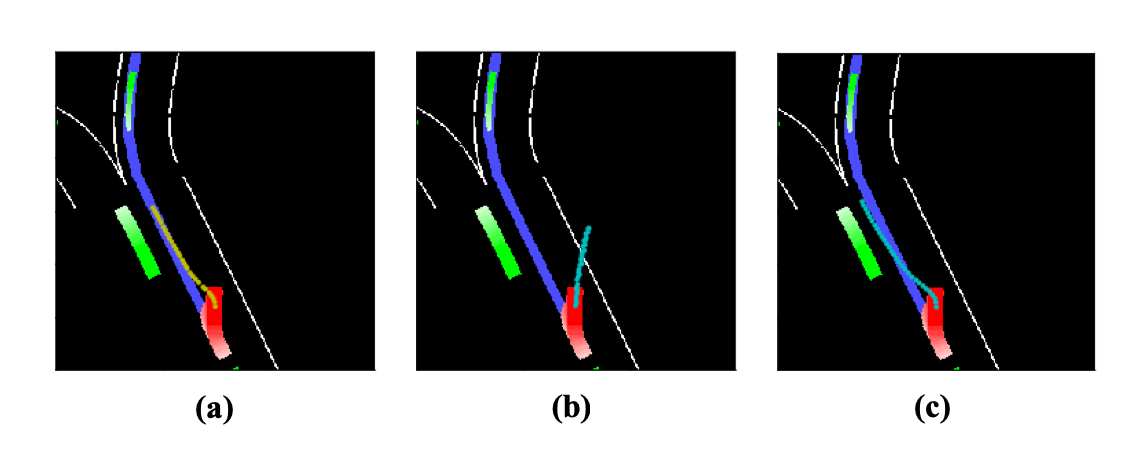} 
	\vspace{-3mm}
  \caption{\em{Comparison between models trained with data augmentation or without data augmentation \textbf{(a)} Groundtruth trajectory \textbf{(b)} Predicted trajectory from $M_0$ \textbf{(c)} Predicted trajectory from $M_1$}}
\label{fig:AugVsNoaug}
\vspace{-7mm}
\end{figure}

We also did an ablation study on how data augmentation improves the performance. We notice that model $M_0$ performs well under most cases. However, once the vehicle runs into abnormal states, $M_0$ can hardly predict a good trajectory to help the vehicle recover to normal states. On the contrary, model $M_1$ has much better ability to help the vehicle recover. Fig.\ref{fig:AugVsNoaug} gives one example.

\begin{figure}
  \centering
    \vspace{-2mm}
	\includegraphics[width=0.5\textwidth]{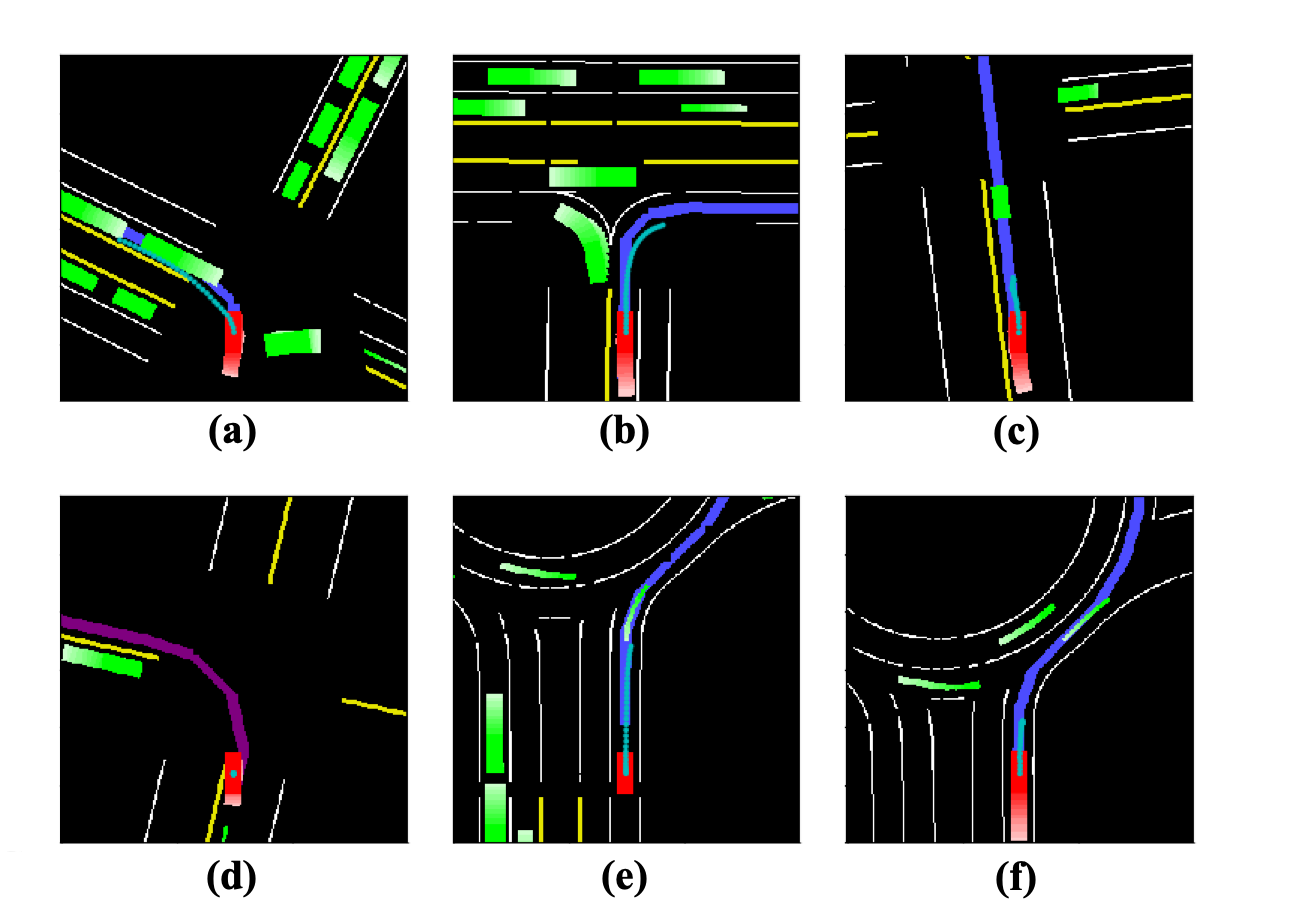} 
	\vspace{-7mm}
  \caption{\em{Example results: \textbf{(a)} Left turn at an intersection. \textbf{(b)} Right turn at an intersection. \textbf{(c)} Blocked by a front vehicle. \textbf{(d)} Stop at red light. \textbf{(e)} \textbf{(f)} Enter a roundabout.}}
\label{fig:Cases}
\vspace{-7mm}
\end{figure}

Moreover, we give several examples of model $M_1$'s outputs. Fig.\ref{fig:Cases} (a) and (b) show that it can output reasonable trajectories in busy intersections. Fig.\ref{fig:Cases} (c) demonstrates that the model learns how to slow down and stop if there is a slow or stopped car in front of the ego vehicle. Fig.\ref{fig:Cases} (d) gives one example that our vehicle stops at the red light. Fig.\ref{fig:Cases} (e) and (f) are examples of entering a roundabout. We can see that the model learns to yield to other vehicles when entering the roundabout in (f).

\subsection{Closed Loop Evaluation}
We also implemented our system in CARLA simulator for closed-loop evaluation. For every 0.1 second, we receive the environment information and render the corresponding bird-view image. The deep neural network policy then performs forward inference and output a predicted trajectory. The trajectory is sent to the tracking controller and then the safety controller to output a control command. The control command is then applied on the ego vehicle in the simulator. This process is repeated until it reaches some terminal criterion. We then evaluate the performance of our models under several urban driving cases and different towns. 

\subsubsection{Evaluation Metrics}
Similar to the metrics designed in~\cite{dosovitskiy2017carla}, we have two metrics for the closed-loop evaluation. The first is success rate, this metric is applied to some specific urban driving cases such as intersection and roundabout. To calculate the success rate, a start and end point are defined for each case, such as start at several meters before entering an intersection/roundabout, and end at several meters after passing through it. Note here we do not report results on simple cases such as lane following as stated in~\cite{dosovitskiy2017carla}, because the model $M_2$ can succeed 100\%. Instead, we perform experiments on two complex cases including a signalized intersection and a roundabout with multiple surrounding dynamic objects. We compare our three models $M_0$, $M_1$ and $M_2$ under the success rate metric.

The second metric is infraction analysis, which we define as the average distance the ego vehicle can run between two collision or out-of-lane events. This definition is a little different with the infraction metric in~\cite{dosovitskiy2017carla}, where they classify collision events with respect to different kinds of objects such as vehicles, bicycles and pedestrians. In this paper, a collision event means collision to any objects. Since we do not have pedestrians in our environment, in order to compare with methods stated in~\cite{dosovitskiy2017carla}, we use the smaller infraction value of their collision-vehicle and collision-bicycle events. This is reasonable because their total collision rate must be higher than that of any single collision type. Our definition for out-of-lane events contains both cases of running to the opposite lane and to the sidewalk, as stated. Similarly, we choose the smaller infraction value to compare. We compare 7 models, including our three models, as well as Modular Pipeline (MP), Conditional Imitation Learning (CIL), Reinforcement Learning (RL) and Conditional Affordance Learning (CAL) shown in~\cite{dosovitskiy2017carla,codevilla2018end,sauer2018conditional}. We also evaluate our performance at a new town (Town01) to see its generalization. Note that we do not evaluate new weather conditions, because our method will not be influenced by different weather because the bird-view representation is not influenced by weather conditions. 

\begin{table}
\centering
\caption{\em{Success rate for the intersection and roundabout scenarios evaluated on our three models $M_0$, $M_1$ and $M_2$. The value represent percentage of success trials}}
\resizebox{0.3\textwidth}{!}{
\begin{tabular}{lccc}
\hline
Task         & $M_0$ & $M_1$ & \multicolumn{1}{c}{$M_2$} \\ \hline
Intersection & 16\%   &  96\%  &      \bf{100\%}                  \\
Roundabout   &  12\%  &  84\%  &    \bf{96\%}                    \\ \hline
\end{tabular}}
\label{success}
\vspace{-5mm}
\end{table}

\begin{table*}
\centering
\caption{\em{Infraction analysis for driving in the training condition (Town03) and new town (Town01) using our three models and other existing methods. The value represents average kilometers traveled between two infractions}}
\resizebox{1\textwidth}{!}{
\begin{tabular}{lcccccccccccccccc}
\hline
                & \multicolumn{8}{c}{Training Condition}                 & \multicolumn{8}{c}{New Town}       \\
Infraction Type & MP & CIL & \multicolumn{1}{c}{RL} & CAL & $M_0$ & $M_1$ & $M_2$ & & & MP & CIL & RL & CAL & $M_0$ & $M_1$ & $M_2$ \\ \hline
Out of lane    &  10.2  &  12.9   &    0.18     &  6.1   &  0.30  &  6.92  & \bf{17.7} & & &  0.45  & 0.76  &  0.23  &  0.88  &  0.29  & 3.77 &  \bf{5.9}  \\
Collision       &  \bf{10.0}  &  3.26  & 0.42   & 2.5 &  0.81  &  3.95  & 8.88 & & &  0.44  & 0.40 & 0.23  & 0.36 &  0.44  &  4.53  &  \bf{11.7}  \\ \hline
\end{tabular}}
\label{infraction}
\vspace{-5mm}
\end{table*}

\subsubsection{Evaluation Results}
Table~\ref{success} shows the success rate for both the intersection and roundabout scenarios of our three models, where we performed 50 trials for each scenarios and each model. We can see that without data augmentation and safety controller, the vehicle can hardly pass these complex urban scenarios as they cannot even make successful turns at a relatively sharp curve road. When trained with augmented data, the success rate improves significantly. Then when safety controller is added, our final model $M_2$ can almost perfectly solve the given scenarios.

Table~\ref{infraction} shows the infraction of our methods and the existing methods on both our training town (Town03) and a new town (Town01), where we performed 50 trials for each model and each town, with 5 minutes for each trial. We then divide the total distance by the total number of infractions to get the infraction value. We can see that our final model $M_2$ outperforms all learning-based methods on both the out-of-lane and collision metrics. The performance of our model $M_2$ is similar to the performance of the modular pipeline under training condition. But for the new town, our model significantly outperforms all other methods. 

Note that our training condition is much more complex than the one in~\cite{dosovitskiy2017carla}, where they train it in Town01 and we train in Town03. Town01 contains only single lane roads with almost no curve roads, and there is no roundabout.

\vspace{-1.5mm}
\subsection{Failure Cases}
\begin{figure}
  \centering
	\includegraphics[width=0.48\textwidth]{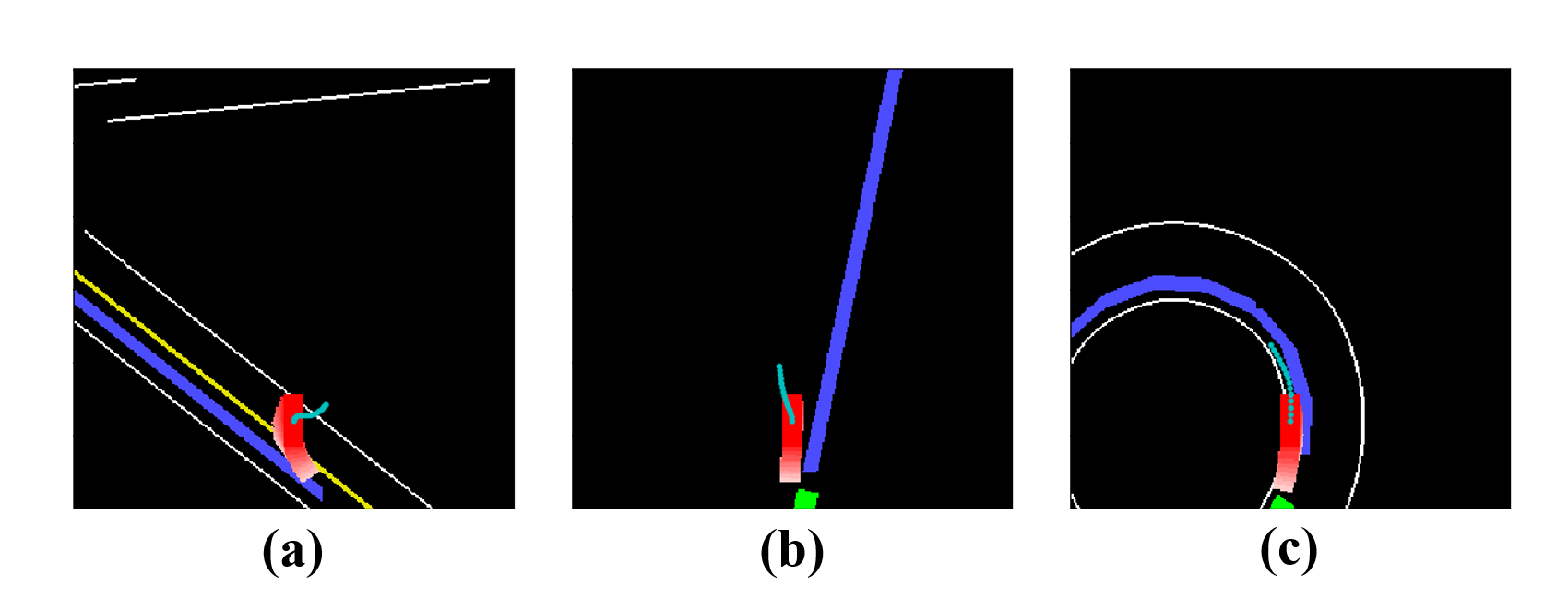} 
	\vspace{-7mm}
  \caption{\em{Failure Cases \textbf{(a)} driving on lane with yellow lane marking on the right \textbf{(b)}  driving on lane with no lane markings \textbf{(c)} driving on roundabout with fence}}
\label{Fig:Failure}
\vspace{-7mm}
\end{figure}

Here we analyze three interesting failure cases we found during our evaluation. Fig.\ref{Fig:Failure}(a) shows a case where the ego vehicle is initialized on a lane where the yellow lane marking is on its right. This makes it look like driving on the opposite lane. As a result, the planned trajectory tries to steer back to the "correct" direction. Although in this case the vehicle fails to follow the given route, the policy has learned something about the structure of the road. Fig.\ref{Fig:Failure}(b) shows a case where there are no lane markings but only the routing information. The vehicle then goes out of lane when there's a fast vehicle behind. Providing more information such as road boundary should help with this situation. Fig.\ref{Fig:Failure}(c) shows a case where the vehicle hits on the fence at a small roundabout. This is because there is no concept of collision in our current model. Reinforcement learning can be incorporated to solve this problem by adding penalties of hitting obstacles.

\vspace{-1.5mm}
\section{CONCLUSION}
In this paper, we proposed and implemented a system to learn a driving policy in generic urban scenarios given offline collected expert driving data, and enhanced the collision avoidance safety. We evaluated our methods on CARLA simulator and found our performance outperform the existing learning-based methods.

In this work we directly get the ground truth information about objects and roads from the simulator, which is impossible in real world. Thus a perception module needs to be developed and the influence of its performance to our system needs to be studied. Furthermore, reinforcement learning methods can be incorporated with our imitation learning model to improve the performance.
\vspace{-2mm}

\bibliographystyle{ieee}
\bibliography{reference}

\end{document}